\documentclass[11pt]{article}

\usepackage[final]{acl}

\usepackage{times}
\usepackage{latexsym}

\usepackage[T1]{fontenc}

\usepackage[utf8]{inputenc}

\usepackage{microtype}

\usepackage{inconsolata}

\usepackage{times}
\usepackage{latexsym}
\usepackage{booktabs}
\usepackage{natbib}
\usepackage{makecell}
\usepackage{hyperref}
\usepackage{tcolorbox}
\usepackage{graphicx}
\usepackage{tikz}
\usepackage{pgfplots}
\usepackage{subcaption}
\usepackage{multirow}
\usepackage[table]{xcolor}
\usepackage[a4paper]{geometry}

\title{RT-SFT: Text Style Transfer from Non-Parallel Corpora by Roundtrip Translation}
\author{Ruoxi Liu \\
  Department of Computer Science \\
 Johns Hopkins University \\
  \texttt{rliu79@jh.edu} \\\And
  Philipp Koehn \\
  Department of Computer Science \\
 Johns Hopkins University \\
  \texttt{phi@jhu.edu} \\}

\begin{document}
\maketitle
\begin{abstract}
Text style transfer (TST) is naturally a supervised task — rewrite a sentence in a target style while preserving its meaning — yet the parallel corpora that supervision requires exist for only a handful of style domains. A common workaround is to \emph{normalize} an input into a style-agnostic intermediate and then \emph{stylize} it into the target style, but the normalizer is typically a lightweight, task-specific paraphraser applied only at test time, feeding a correspondingly small stylizer. We observe that a style-stripping normalizer already exists at scale: neural MT systems trained on hundreds of millions of general-domain sentence pairs preserve content while regressing toward generic phrasing, so roundtrip translation through a pivot language strips stylistic signal without any task-specific training. This turns normalization from an inference-time patch into a data-generation tool. Roundtrip-translating a monolingual in-style corpus yields a pseudo-parallel corpus on which we LoRA-finetune an instruction-tuned LLM as the stylizer (RT-SFT); applying the same normalizer to test queries keeps that stylizer in-distribution. We show that across four style domains, RT-SFT outperforms state-of-the-art methods, such as few-shot in-context learning, by considerable margins. We also report on effective retrieval augmentation methods for expert style domains with strict terminology and naming conventions.
\end{abstract}

\section{Introduction}

Text Style Transfer (TST) is the task of rephrasing text by modifying stylistic attributes while preserving its core, attribute-independent semantics and intent \cite{Shen2017,Toshevska2024}. These attributes encompass formality, attitude, verbosity, preferred terminology, and other characteristics inherent to the text. A central obstacle in TST is the scarcity of annotated parallel corpora, which prevents the use of fully supervised training or finetuning \cite{Pan2024} in most style domains.

One established approach to this problem is \emph{normalize-then-stylize}: a normalizer maps an input to a style-agnostic form, either a latent encoding or text, and a stylizer regenerates that form in the target style. STRAP \citep{Krishna2020}, for instance, normalizes by paraphrasing the input to strip its style before decoding with a style-specific generator. The advantage of this framing is that the normalizer defines a single shared domain through which every input passes, whatever style it arrives in, so the stylizer never has to accommodate unpredictable source styles.

Robustness to diverse, unpredictable input styles therefore depends on the training data for the normalizer. Most current normalizers are trained on curated \emph{paraphrase corpora} such as ParaFusion \citep{ParaFusion2024}. Such corpora are reported to offer limited lexical and syntactic diversity \citep{ParaBank2,Krishna2020,ParaFusion2024}, and paraphrasers trained on them degrade on stylistic features unseen during training. A second line of work prompts an instruction-tuned LLM to paraphrase without any parameter update \citep{ConversationStyle2023,SynDec2025}. We argue that this is less principled: the stylizer unavoidably carries the stylistic features and biases of the prompted LLM.

\begin{figure}
    \centering
    \includegraphics[width=1\linewidth]{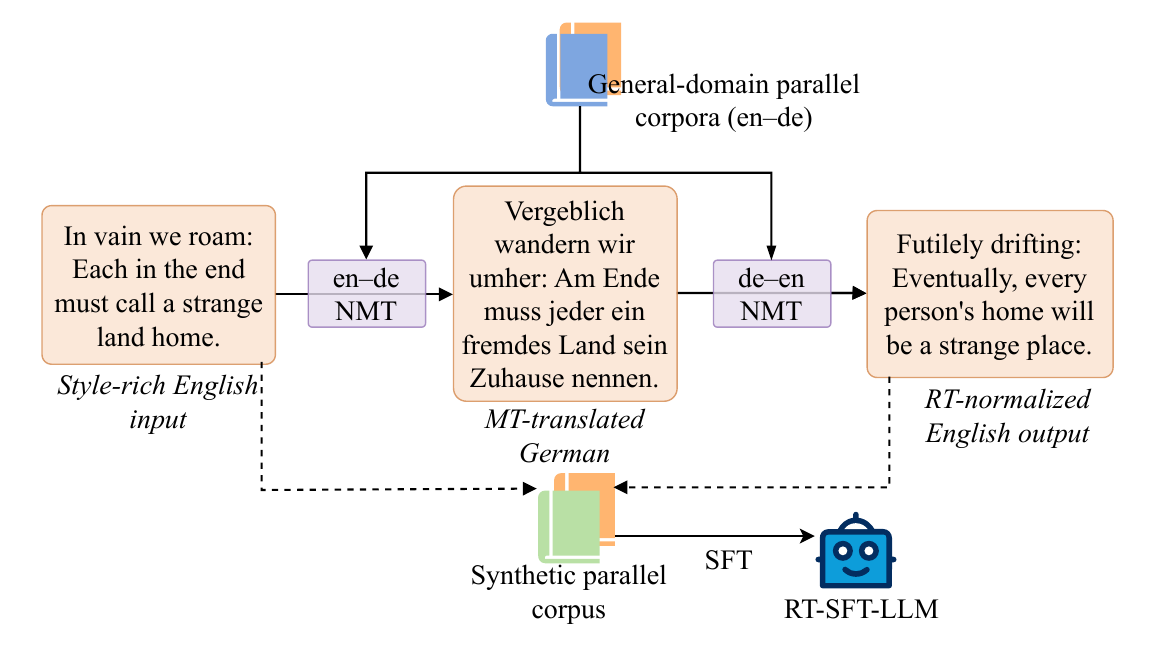}
    \caption{Our proposed workflow for finetuning large language models (LLMs) for text style transfer (TST) using only non-parallel data in the target domain. A bilingual general-domain parallel dataset is used to train a pair of neural machine translation (NMT) models capable of translating between English and a pivot language. We then obtain RT-normalized texts of the original in-domain texts by roundtrip translating the in-domain set with the NMT models. This enables supervised finetuning of LLMs for TST, where we finetune LLMs for RT-normalized-domain to target-domain transfer using the synthetic parallel corpus. }
    \label{fig:RT}
\end{figure}

We propose roundtrip translation as the normalizer: a sentence is translated into a pivot language and then back into the original language, a technique previously used to assess MT robustness and generation quality \citep{somers2005round,DBLP:journals/corr/abs-2004-13937}. This requires neither paraphrase supervision nor a prompted LLM (Figure~\ref{fig:RT}). We first train two Marian NMT models \citep{Marcin2018} on general-domain parallel corpora \citep{kocmi-etal-2025-findings}, and compose them into an English--pivot--English roundtrip pipeline. Applying this pipeline to a non-parallel corpus in the target style produces an RT-normalized counterpart for each sentence, and the two sides form a pseudo-parallel corpus that restores a supervised TST setting in a domain with no bitext. We then finetune an instruction-tuned LLM on this corpus as the stylizer (an RT-SFT-LLM), with retrieval augmentation applied to both the finetuning and the inference prompts (\textsection\ref{sec:method_RAG}). At test time we pass the query through the same roundtrip pipeline before generation, so that the stylizer receives the RT-normalized input it was trained on (\textsection\ref{sec:method_RT}). We show in \textsection\ref{sec:exp_inference} that this alignment raises style accuracy in all four domains.

We evaluate our method (\textsection\ref{sec:exp}) on four style domains with distinctive stylistic features (\textsection\ref{sec:exp_setup_data}), against two established baselines: few-shot In-Context Learning (ICL) \citep{Reif2021,Suzgun2022} and Automatic Post-Editing (APE) \citep{Simard2007, 9714400}. Following prior work, we measure transfer quality with BERT-based style classifiers trained on held-out data and content preservation with BLEU \citep{Subramanian2018,Wan2023,Aycock2024}.

Our \textbf{main contributions} are:

\begin{itemize}
    \item \textbf{Pseudo-parallel dataset construction} (\textsection\ref{sec:method_RT}). We use roundtrip translation through general-domain MT as a style normalizer, synthesizing the parallel corpus that supervised finetuning requires in domains that lack bitext.
    \item \textbf{Retrieval augmentation for finetuning and inference} (\textsection\ref{sec:method_RAG}). We apply retrieval augmentation to the \textbf{finetuning} prompts as well as the inference prompts, and show that its benefits extend beyond prompting by aligning the retriever with the generator.
    \item \textbf{Systematic evaluation of RT-SFT-LLMs} (\textsection\ref{sec:exp}). We benchmark RT-SFT-LLMs across four style domains using four LLMs as base model, three prompt templates, and several retrieval and inference configurations, against established baselines.
\end{itemize}

\section{Related Work}

\paragraph{Unsupervised and semi-supervised TST} Because corpora that align the same content across styles are scarce, most work performs style transfer from non-parallel data. Early neural approaches disentangle content from style in a latent space and recombine the content with a target-style representation \cite{Shen2017,Nangi2021,Voigt2018}, typically using adversarial objectives and auxiliary losses that preserve content through the transfer \cite{Tian2018,Nouri2022}. Semi-supervised methods further reduce the reliance on task-specific data by exploiting generic pretrained resources and reward models \cite{lai-etal-2021-thank}. More recent work abandons explicit disentanglement, instead synthesizing pseudo-parallel data with LLMs and refining generation with reinforcement learning: \citet{Pan2024} couple LLM prompting with attention masking, while intensity-controllable methods first supervise-finetune a base transfer model and then apply reinforcement learning with hierarchical rewards to modulate stylistic strength \cite{ControllableIntensity2026}. Our method also assumes only non-parallel, target-style data, but recovers a supervised training signal through normalization (\textsection\ref{sec:methods}).

\paragraph{Normalize-then-stylize} A strategy for non-parallel TST is to first map the input to a style-agnostic intermediate---to \emph{normalize} it---and then generate in the target style---to \emph{stylize} it. The normalizer has taken many forms: a learned diverse paraphraser \cite{Krishna2020}, roundtrip or back-translation through a pivot language whose output is assumed to retain content but not style \cite{Prabhumoye2018,rabinovich-etal-2017-personalized,Zhang2018,He2020}, denoising and latent-variable encoders \cite{Subramanian2018,Riley2021}, abstract-meaning-representation graphs \cite{Jangra2022}, and, most recently, diffusion paraphrasers \cite{Horvitz2024}. In these methods the normalizer is typically applied \emph{at inference time} and paired with a comparatively small stylizer: STRAP \cite{Krishna2020} paraphrases the input at test time to strip its style and then decodes with a style-specific GPT-2; \citet{TinyStyler2024} reconstruct text from a neutralized paraphrase conditioned on an authorship embedding; conversational methods reduce an utterance to a style-free form and rewrite it \cite{ConversationStyle2023}; and synthesize-then-decode prompting amplifies the target style only at decoding \cite{SynDec2025}. We adopt this paradigm but realize it with two components trained at scale---a general-domain neural machine translation (NMT) model as the normalizer (through roundtrip translation) and an instruction-tuned LLM, finetuned to map normalized text to the target style, as the stylizer---together with retrieval augmentation at both training and inference (\textsection\ref{sec:methods}). This trained, LLM-scale realization is precisely what prior, largely inference-time methods lack.

\paragraph{Prompt engineering and in-context learning} A parallel line of work performs TST with LLMs purely through prompting, without updating any parameters. \citet{Reif2021} frame transfer as instruction-guided rewriting with augmented zero-shot prompts; other methods rerank sampled candidates by style strength and fluency \cite{Suzgun2022}, supply retrieved in-domain exemplars as few-shot ``shots'' \cite{Liu2024c,Chen2024b,Bhandarkar2024,Aycock2024}, or optimize the prompt template and its retrieved context \cite{Zhang2024,Wan2023,Tao2024}. The most recent methods steer generation through register analysis \cite{RegisterSteering2025}, structure long-document transfer \cite{LongTextTST2025}, and probe how far prompting alone can go \cite{IEEEStepForward2025,Mukherjee2024}. Such approaches require no training, but they are inherently prompt-bound: they cannot adapt model parameters to a target domain, which limits terminology control and long-range stylistic consistency. We instead finetune the stylizer (\textsection\ref{sec:methods}).

\begin{figure}
    \centering
    \includegraphics[width=0.8\linewidth]{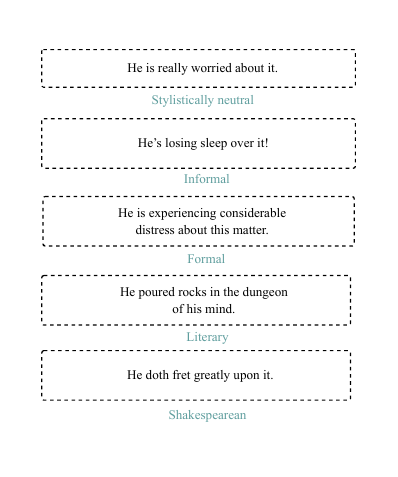}
    \caption{ Example sentences illustrating semantically equivalent content in various styles. Outputs of our roundtrip translation pipeline are considered RT-normalized.}
    \label{fig:domain}
\end{figure}

\paragraph{Supervised TST and parameter-efficient finetuning} When aligned data are available, style transfer can be cast as supervised sequence-to-sequence learning, as in \citet{jhamtani-etal-2017-shakespearizing}'s copy-augmented model for Shakespearean English. The bottleneck is data: parallel style corpora exist for only a handful of domains---modern-to-Shakespearean English \cite{Xu2012}, formality \cite{rao-tetreault-2018-dear,Briakou2021}, biblical prose \cite{Carlson2018}, expert-versus-layman medical text \cite{Cao2020}, and, more recently, parallel detoxification corpora \cite{MultiParaDetox2025,NineLangDetox2025}---and new benchmarks continue to target these same well-resourced settings \cite{TSTBench2025,StylePersonalizedEval2025}. Parameter-efficient finetuning of LLMs for TST has begun to appear \cite{Liu2024a,Mukherjee2024} but remains confined to domains with existing parallel data. We remove this constraint: by normalizing a monolingual target-style corpus, we synthesize the parallel data that supervised finetuning requires and extend it to domains with no bitext (\textsection\ref{sec:methods}).

\section{Methods}\label{sec:methods}

\subsection{Roundtrip Translation}\label{sec:method_RT}

\begin{figure*}
    \centering
    \includegraphics[width=1\linewidth]{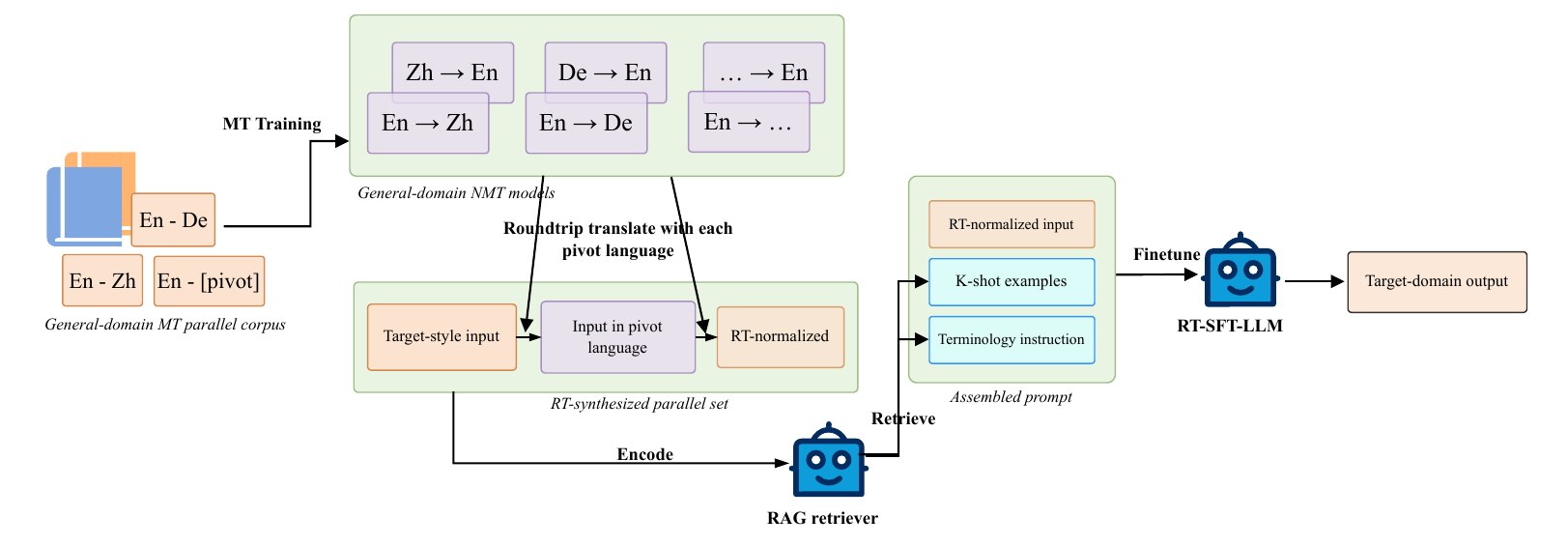}
    \caption{\textbf{Our proposed workflow.} We train MT systems using general domain parallel data for Chinese-English and German-English two-way translations, and roundtrip-translate target style text to obtain RT-normalized counterparts. We roundtrip translate every record with both pivot languages and mix the synthetic corpus. The synthetic parallel corpus is then used for finetuning a stylizer LLM, with retrieved information injected in the prompt.}
    \label{fig:overview}
\end{figure*}

Unlike most state-of-the-art methods that instantiate the \emph{normalize-then-stylize} paradigm \cite{Krishna2020}, we train both components at scale. The normalizer is a pair of neural machine translation (NMT) models, trained with Marian \cite{Marcin2018} on a large general-domain corpus between English and a pivot language; composed into a roundtrip translation pipeline, it strips pronounced stylistic features while preserving content, yielding \emph{RT-normalized} text; Figure~\ref{fig:domain} shows RT-normalized outputs alongside the same content rendered in each target style. Applying it to a monolingual in-style corpus pairs every sentence with its RT-normalized counterpart, producing the pseudo-parallel corpus on which we parameter-efficiently finetune an instruction-tuned LLM as the stylizer (an \emph{RT-SFT-LLM}; Figure~\ref{fig:RT}). At test time the two run in sequence, the normalizer followed by the stylizer (Figure~\ref{fig:overview}).

This sequencing resolves a distribution mismatch: during finetuning the stylizer sees only RT-normalized inputs, whereas an arbitrary query at test time may carry pronounced, unseen style. Passing the query through the same roundtrip pipeline first mirrors the training-time construction exactly, and is the analogue of STRAP's inference-time paraphrasing \cite{Krishna2020}, realized here by an NMT normalizer. We compare RT-first inference against feeding the raw query directly in \textsection\ref{sec:exp_inference}, and find that it improves target-style conformity, especially for stylistically diverse or out-of-domain inputs.

\subsection{RT-SFT with Parameter Efficient LoRA Finetuning}\label{sec:method_lora}
LoRA (Low-Rank Adaptation of Large Language Models) is an efficient approach that reduces computational and memory costs by using low-rank approximation techniques \cite{hu2021lora}. The LoRA approach involves freezing the pre-trained model's weight matrices and introducing trainable low-rank decomposition matrices into the model's layers. This approach allows us to finetune the 7B and 8B LLMs with 2 NVIDIA A100 GPUs, each with 80GB of memory. Hyperparameters and configurations we used are given in Appendix~\ref{sec:appendix_config}.

\subsection{RAG Retrievers for RT-SFT-LLMs}\label{sec:method_RAG}

\citet{lewis_retrieval-augmented_2020} proposed the Retrieval Augmented Generation (RAG) framework in which a retriever-generator model is trained end-to-end to enhance coherence between the pre-trained retriever and generator subsystems. In a similar spirit, our RAG system injects retrieved information into both finetuning and inference-time prompts for RT-SFT-LLM, unlike previous RAG-based TST methods where RAG is primarily considered a prompting technique \cite{Liu2024c,Wan2023,Aycock2024,Zhang2024}. In \textsection\ref{sec:exp_RAG}, we present the performance of our system with increasing levels of retrieval augmentation compared against our method without RAG (RT-SFT 0-shot) and state-of-the-art inference-time RAG without training (5-shot In-Context Learning). We show that RAG introduced at training time produces considerable improvements compared to both RT-SFT without RAG and inference-time RAG.

\begin{figure*}
    \centering
    \includegraphics[width=1\linewidth]{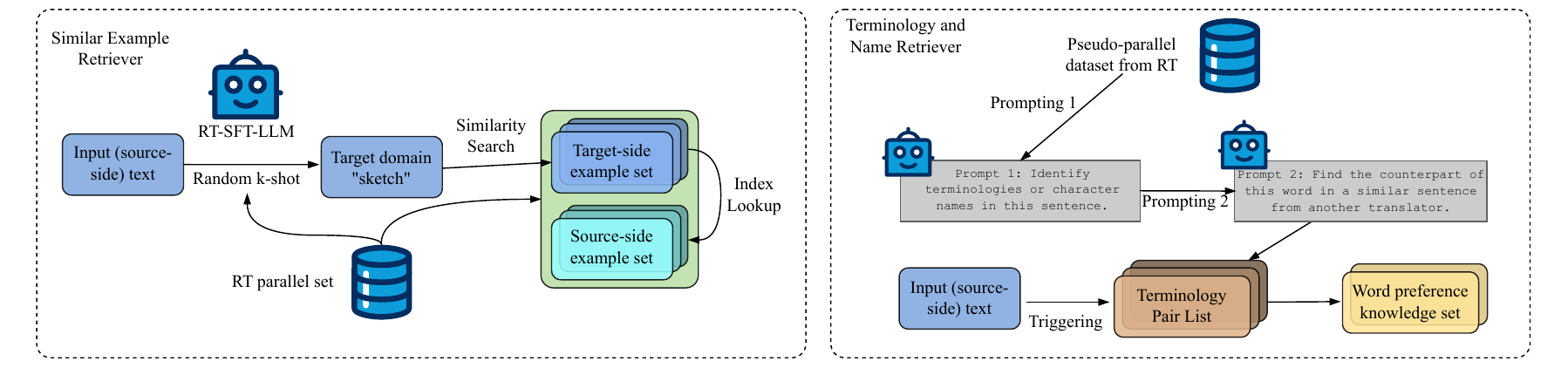}
    \caption{\textbf{Retrieval augmentation workflow. }\textbf{Left (a):} Relevant example retriever. We encode \textbf{the target-side texts} of the parallel synthetic datasets for nearest-neighbor search. For each query, we first do k-shot inference with the RT-SFT-LLM to obtain an ``in-domain'' sketch, which is used as search query in the target-side dataset to obtain the k most relevant pairs. \textbf{Right (b)} Terminology and name retriever: For each instance in the synthetic parallel datasets, the first LLM call extracts relevant words from the source side, then the second call matches them with their counterparts in the target side, yielding a terminology pair list for each domain. }
    \label{fig:RAG}
\end{figure*}

\subsubsection{Relevant Example Retrieval Augmentation}\label{sec:method_RAG_ex_train}

The first type of retrieval augmentation we report on obtains sentence pairs as instructions for how we would like the query to be transferred. Results demonstrate that examples relevant to the query are more instructive and bring stronger performance improvements (\textsection\ref{sec:exp_RAG}).

We vectorize the pseudo-parallel corpus and use Faiss \cite{douze_faiss_2024} (Figure~\ref{fig:RAG}) to retrieve nearest neighbors, following the original RAG schema \citep{lewis_retrieval-augmented_2020}. Interestingly, we found that searching with the target side (in-domain) retrieves better instances than searching with the source (out-of-domain) side, giving slightly better results. We thus use retrieved examples with relevant target sides for example prompt injection. Specifically, we exclude the current instance itself during retrieval to avoid leakage.

For inference-time RAG where the target side text is unavailable and is what we want to predict, we take a ``sketch-first'' approach following \citet{Wang2022}'s two-round retrieval schema (Figure~\ref{fig:RAG}). We first perform few-shot inference with \textbf{randomly selected} examples to generate a draft ``sketch'' output that resembles the in-domain transferred generation, though of limited quality due to the random shots. We then use this draft as the query to retrieve the most relevant examples from the Faiss vector bank for a refined second-round inference. In \textsection\ref{sec:exp_inference}, we show that relevant example augmentation at inference time through this approach yields better performance than randomly selected examples and inference without any examples (Table~\ref{tab:term_comparison}). 

\subsubsection{Terminology and Name List Retrieval}\label{sec:method_RAG_term}

Diction and word preference are important aspects of text style domains. The same concept or object can be referred to by different terminologies in different domains. In literary translation domains, there is a similar issue of \textbf{naming consistency}, where machine-translated works may use semantic translations and direct translations in different contexts to refer to the same characters, causing confusion \cite{matusov-2019-challenges}.

We improve our RT-SFT-LLM's terminology correctness and long-context consistency by retrieving a \textbf{terminology and name list} from our pseudo-parallel corpus (Figure~\ref{fig:RAG}). For each record in the pseudo-parallel corpus, we first prompt an LLM with the source side of the paired sentences and ask it to identify any domain-specific terms or names in it. Then, we do a second round of prompting with the target side of the sentence pairs alongside the retrieved domain-specific terms, and ask the LLM to find their counterparts. Through this pipeline, we construct a list of source domain to target domain preferred terminology pairs. If any of the source-side words are present in the query, we add a one-sentence instruction in the inference prompt that provides terminology and name transfer guidance. The prompts we used are in Appendix~\ref{sec:appendix_prompt}.

\subsection{Evaluation: BLEU and Style Classification Accuracy Score}\label{sec:method_acc}

We primarily evaluate two aspects of our models, namely style-transfer quality and content preservation. For each style domain we train a BERT-based \cite{47751} style classifier on held-out in-domain data, following \citet{Liu2024a, Liu2024c, Mukherjee2024}. The classifier scores a given text between 0 and 1, where we treat 1 as "in-domain" and 0 as "out-of-domain". We apply the trained classifier to the output of our RT-SFT-LLMs to yield a style-classification accuracy that measures how well the output matches the target style. BLEU is computed between the generation and the reference in-domain text, measuring how well the transferred output preserves the intended content.

\section{Experiments}\label{sec:exp}

\subsection{Experiment Setup}\label{sec:exp_setup}

\subsubsection{Datasets, Synthetic Data Generation, and Baselines}\label{sec:exp_setup_data}

\begin{table}[ht]
\centering
\resizebox{\columnwidth}{!}{%
\begin{tabular}{l r r r}
    \toprule
    \textbf{Dataset} & \textbf{Training} & \textbf{Dev} & \textbf{Test} \\
    \midrule
    WMT26-eng-zho & \underline{241,429,382} & \underline{3,000} & -- \\
    WMT26-deu-eng & \underline{1,128,176,715} & \underline{3,000} & -- \\
    \hline
    MSD          & 130,199 & \underline{150}   & \underline{675}   \\
    Shakespeare  & 24,852  & \underline{1,313} & \underline{1,293} \\
    GYAFC        & 103,562 & \underline{1,000} & \underline{2,748} \\
    ParaDetox    & 19,466  & \underline{150}   & \underline{150}   \\
    \bottomrule
\end{tabular}
}
\caption{%
  \textbf{Datasets.} \underline{Underlined} counts are parallel (aligned) sentence pairs; plain counts are monolingual sentences. The WMT26 General Machine Translation corpora \citep{kocmi-etal-2025-findings} are used to train the roundtrip-translation normalizers; we use only the parallel portion of the released splits and manually hold out 3{,}000 sentence pairs per language pair as dev set. The four style datasets are used as training data for our RT-SFT method. Note that although 3 of the 4 style domains provide parallel training data, we only use the in-domain side of them during training, as our system only requires in-domain non-parallel training data. We use the provided, human-annotated parallel sentences only for the dev and test sets.
}
\label{tab:4col}
\end{table}

\textbf{Machine-translation data} We build the roundtrip translation normalizer from two general-domain parallel corpora released for the WMT26 General Machine Translation shared task \citep{kocmi-etal-2025-findings}, covering English--Chinese and German--English. We use only the parallel portion of the released splits to train the Marian NMT models \citep{Marcin2018}, holding out 3{,}000 sentence pairs per language pair as a development set; detailed configurations are given in Appendix~\ref{sec:appendix_config}. Every in-style sentence is roundtrip translated through both the Chinese and the German pivot, and the two resulting RT-normalized--to--target pairs are pooled into a single synthetic corpus for finetuning. Corpus sizes are reported in Table~\ref{tab:4col}.

\textbf{Style-transfer datasets.} We evaluate on four publicly available benchmarks, each targeting a specific style and providing a human-aligned parallel test set. The domains are: expert medical text (\textbf{MSD}; \citealp{Cao2020}), Shakespearean English (\textbf{Shakespeare}; \citealp{Xu2012}), formal text (\textbf{GYAFC}; \citealp{rao-tetreault-2018-dear}), and detoxified text (\textbf{ParaDetox}; \citealp{Logacheva2022}). Split sizes are listed in Table~\ref{tab:4col}, where underlined counts denote parallel data and plain counts denote monolingual data.

\textbf{Baselines: }First, we set up an \textbf{In-Context Learning (ICL)} baseline method by prompting the LLMs with the same prompt from our own method, with the addition of few-shot examples retrieved based on similarity and the absence of finetuning. The second baseline we adopt is \textbf{Automatic Post-Editing (APE)}: we train an additional MT model for each domain on the output (RT-normalized) versus text input (target style), so that it performs English-to-English ``translations'' that bring RT-normalized texts to in-domain. This is a widely-used practice that accompanies Machine Translation systems \citep{Simard2007, 9714400}, where the additional APE module learns to correct systematic errors in the MT system through NMT training. Our variant is monolingual (source-free), post-editing the roundtrip output directly to match the target text, which is the input to the roundtrip translation pipeline. 

\subsubsection{TST Prompt Templates}\label{sec:exp_setup_prompt}
We experimented with three potential prompt templates for TST finetuning. Prompt details and experiments on prompts are in Appendix~\ref{sec:appendix_prompt}. After testing and careful evaluation, we decided to use the following prompt template for our method:

\begin{figure}[htbp]
\centering
\begin{tcolorbox}[colback=gray!10, colframe=black, width=\columnwidth, sharp corners]
\small
\textbf{Rewrite the given sentence into the style of [style name].} \\
Here are [n] examples: \\
Input: \textcolor{red}{[example input i]}. Output: \textcolor{red}{[example output i]}. ......\\
Note that you may want to rewrite "\textcolor{red}{[input term]}" to "\textcolor{red}{[output term]}" for contextual consistency. \\
Now go ahead: Input: \textcolor{blue}{[query input]}. The [style name] output:
\end{tcolorbox}
\label{prompt}
\end{figure}

\subsection{Overall Performance Comparison}\label{sec:exp_RAG}

\begin{table*}[h]
    \centering
    \resizebox{\textwidth}{!}{%
    \begin{tabular}{@{}p{5.5cm}cccccccc@{}} 
        \toprule
        \multirow{2}{*}{\textbf{Method}} & \multicolumn{2}{c}{\textbf{MSD}} & \multicolumn{2}{c}{\textbf{Shakespeare}} & \multicolumn{2}{c}{\textbf{GYAFC}} & \multicolumn{2}{c}{\textbf{ParaDetox}} \\
        \cmidrule(lr){2-3} \cmidrule(lr){4-5} \cmidrule(lr){6-7} \cmidrule(lr){8-9}
        & \textbf{Acc.} & \textbf{BLEU-ref} & \textbf{Acc.} & \textbf{BLEU-ref} & \textbf{Acc.} & \textbf{BLEU-ref} & \textbf{Acc.} & \textbf{BLEU-ref} \\
        \midrule
        COPY & 0.107 & 14.01 & 0.009 & 15.35 & 0.149 & 16.54 & 0.012 & 20.03 \\
        5-shot In-Context Learning~\citeyearpar{Reif2021} & 0.396 & 9.58 & 0.771 & 12.79 & 0.521 & 19.34 & 0.792 & \textbf{21.35} \\
        Automatic Post-Editing~\citeyearpar{9714400} & 0.672 & 12.21 & 0.693 & 11.90 & 0.435 & 17.74 & 0.720 & 17.35 \\
        \hline
        RT-SFT 0-shot & 0.706 & 11.32 & 0.841 & 11.47 & 0.591 & 21.33 & 0.820 & 18.40 \\
        RT-SFT random-3-shot & 0.715 & 13.35 & 0.809 & 13.21 & 0.620 & 22.41 & 0.893 & 18.11 \\
        RT-SFT random-5-shot & 0.732 & 13.61 & 0.794 & 13.42 & 0.611 & 22.22 & 0.891 & 18.09 \\
        RT-SFT relevant-3-shot & \textbf{0.797} & 13.59 & 0.841 & 14.65 & 0.642 & \textbf{24.79} & \textbf{0.935} & 17.91 \\
        RT-SFT relevant-5-shot & 0.782 & 13.53 & 0.839 & 14.77 & \textbf{0.644} & 24.32 & 0.931 & 19.35 \\
        \quad + terminology and name retrieval & 0.763 & \textbf{13.79} & \textbf{0.842} & \textbf{17.35} & 0.612 & 24.17 & 0.912 & 19.91 \\
        \bottomrule
    \end{tabular}%
    }
    \caption{\textbf{Main results.} TST performance of RT-SFT against baselines across the four style domains (base model: Llama-3.1-8B-Instruct; synthetic corpus pooled from German and Chinese roundtrip pivots). \textit{random} and \textit{relevant} indicate whether the $k$ in-context example pairs are sampled at random or retrieved for relevance; \textit{+ terminology and name retrieval} adds domain-specific term guidance when triggered (\textsection\ref{sec:method_RAG_term}). Best result per column in bold, excluding COPY: returning the input unchanged attains very high BLEU at near-zero style accuracy, a known artefact of BLEU-based TST evaluation \citep{Pang2019}. RT-SFT improves style accuracy over both baselines in all four domains and in every configuration tested, and relevant-example retrieval improves accuracy over random examples in all eight paired comparisons.}
    \label{tab:exp_RAG}
\end{table*}

This subsection compares roundtrip-translation supervised finetuning (\textbf{RT-SFT}) against three reference points across the four style domains (Table~\ref{tab:exp_RAG}). We use Llama-3.1-8B-Instruct as the base model and are finetuned on the synthetic corpus of \textsection\ref{sec:exp_setup_data}, which pools roundtrip translations through both pivots (Chinese and German); an experiment on various LLMs as the base model is reported in \textsection\ref{sec:exp_basellm}. \textbf{COPY} offers a calibration where we measure the BLEU and accuracy of the input itself against the target text. Among RT-SFT variants, \textbf{$k$-shot (random)} supplies $k$ randomly drawn RT-normalized-to-target pairs and \textbf{$k$-shot (relevant)} the $k$ most relevant pairs, retrieved by cosine-distance search at finetuning time and by the procedure of \textsection\ref{sec:method_RAG_ex_train}; the final variant adds \textbf{terminology and name retrieval} (\textsection\ref{sec:method_RAG_term}). Results are single runs, and we do not interpret accuracy differences below $0.02$.

Style accuracy is measured by a BERT classifier trained per domain (\textsection\ref{sec:method_acc}), so the COPY row calibrates it: it reports how often an untransferred input is already labelled in-domain. The near-zero values on Shakespeare ($0.009$) and ParaDetox ($0.012$) show that these boundaries are almost linearly separable, whereas MSD ($0.107$) and GYAFC ($0.149$) are harder to classify, and their scores must be read against a higher floor and a correspondingly lower ceiling ($0.797$ and $0.644$, against $0.842$ and $0.935$). Against this calibration, RT-SFT improves accuracy over both baselines in all four domains and in every configuration tested---24 of 24 cells. Neither half of the method suffices alone: zero-shot RT-SFT, which uses no retrieval at all, beats ICL by $+0.310$, $+0.070$, $+0.070$ and $+0.028$, and beats APE by $+0.034$, $+0.148$, $+0.156$ and $+0.100$. Retrieval augmentation therefore adds to these gains rather than producing them. APE leads ICL only on MSD ($0.672$ vs.\ $0.396$), the domain with the most specialized lexicon.

\begin{table*}[h]
    \centering
    \resizebox{\textwidth}{!}{%
    \begin{tabular}{@{}lcccccccc@{}}
        \toprule
        \multirow{2}{*}{\textbf{Inference methods}} & \multicolumn{2}{c}{\textbf{MSD}} & \multicolumn{2}{c}{\textbf{Shakespeare}} & \multicolumn{2}{c}{\textbf{GYAFC}} & \multicolumn{2}{c}{\textbf{ParaDetox}} \\
        \cmidrule(lr){2-3} \cmidrule(lr){4-5} \cmidrule(lr){6-7} \cmidrule(lr){8-9}
        & \textbf{Acc.} & \textbf{BLEU-ref} & \textbf{Acc.} & \textbf{BLEU-ref} & \textbf{Acc.} & \textbf{BLEU-ref} & \textbf{Acc.} & \textbf{BLEU-ref} \\
        \midrule
        COPY & 0.107 & 14.01 & 0.009 & 15.35 & 0.149 & 16.54 & 0.012 & 20.03 \\
        5-shot In-Context Learning~\citeyearpar{Reif2021} & 0.396 & 9.58 & 0.771 & 12.79 & 0.521 & 19.34 & 0.792 & \textbf{21.35} \\
        Automatic Post-Editing~\citeyearpar{9714400} & 0.672 & 12.21 & 0.693 & 11.90 & 0.435 & 17.74 & 0.720 & 17.35 \\
        \hline
        0-shot inference & 0.730 & 13.31  & 0.791 & 17.21 & 0.595 & 23.96 & 0.909 & 17.07 \\
        random 5-shot inference & 0.759 & 11.53 & 0.841 & 17.01 & 0.575 & 21.17 & 0.893 & 18.63 \\
        relevant 5-shot inference without RT normalization & 0.671 & \textbf{14.55} & 0.731 & \textbf{18.69} & 0.556 & \textbf{26.62} & 0.902 & 20.21 \\
        relevant 5-shot inference & \textbf{0.763} & 13.79 & \textbf{0.842} & 17.35 & \textbf{0.612} & 24.17 & \textbf{0.912} & 19.91 \\
        \bottomrule
    \end{tabular}%
    }
    \caption{TST performance under different inference procedures. All groups use one Llama-3.1-8B-Instruct RT-SFT-LLM, finetuned with 5-shot relevant examples and terminology retrieval (\textsection\ref{sec:exp_RAG}); baseline rows are repeated from Table~\ref{tab:exp_RAG} for reference. All RT-SFT rows apply RT-first inference except \textit{without RT normalization}, which feeds the raw query. Best result per column in bold, excluding COPY. RT-first inference raises style accuracy and lowers BLEU in all four domains; relevant-example retrieval raises BLEU in all four while leaving accuracy largely unchanged.}
    \label{tab:term_comparison}
\end{table*}

BLEU must be read jointly with accuracy, since the two trade off systematically in TST: it is widely reported by prior work that untransferred input routinely outscores most systems \citep{Pang2019, Mukherjee2023}. Our COPY row reproduces this---considerably high BLEU across all four domains, and the highest BLEU on MSD ($14.01$) at only $0.107$ accuracy---and explains the single column in which a baseline leads: 5-shot ICL takes the top ParaDetox BLEU ($21.35$) at a comparatively low style accuracy ($0.792$, against $0.912$--$0.935$ for the relevant-shot RT-SFT variants); consistent with the style--content trade-off, it under-edits toward the input. A BLEU drop relative to COPY is thus expected wherever source and reference already overlap heavily. Where they genuinely diverge the trade-off does not bind: on GYAFC and Shakespeare the best RT-SFT variant leads on both metrics, improving BLEU by $+5.45$ and $+4.56$ over the stronger baseline.

Retrieval improves both metrics, but unevenly. Relevant examples beat random ones on accuracy in all eight paired comparisons (mean $+0.043$) and on BLEU in six of eight (mean $+1.06$), with gains concentrated where a length- and vocabulary-matched exemplar is available: $+2.38$ on GYAFC and $+1.44$ on Shakespeare at $k{=}3$, against $+0.24$ on MSD. The number of shots has no consistent effect, 3-shot winning five of eight cells by margins well below the relevant-vs-random gap. Terminology and name retrieval behaves as the lexical intervention it is: added to relevant-5-shot it raises BLEU in three of four domains---most on Shakespeare ($+2.58$, the largest single gain in the table), where the style is carried by a fixed vocabulary and by character names---but on GYAFC, where formality is realized syntactically, it lowers both metrics. Averaged over the four domains it trades $0.017$ accuracy for $0.82$ BLEU. We therefore recommend it for domains with a stable terminology or naming convention, and relevant-example retrieval alone elsewhere.

\subsection{Inference-time methods}\label{sec:exp_inference}

We next perform ablation studies on the inference procedure (Table~\ref{tab:term_comparison}). All groups use the same Llama-3.1-8B-Instruct RT-SFT-LLM, finetuned with 5-shot relevant examples and terminology retrieval (\textsection\ref{sec:exp_RAG}), and differ only in what is done at test time: \textbf{0-shot} prompts with the task description alone, while \textbf{random} and \textbf{relevant 5-shot} append examples sampled at random or retrieved by the procedure of \textsection\ref{sec:method_RAG_ex_train}. The \textbf{without RT normalization} row feeds the raw query directly into the finetuned stylizer LLM without being normalized by the RT pipeline, and provides the controlled contrast isolating RT-first inference.

RT-first inference yields a strictly monotone trade: style accuracy rises in all four domains ($+0.092$, $+0.111$, $+0.056$, $+0.010$) and BLEU falls in all four ($-0.76$, $-1.34$, $-2.45$, $-0.30$). Aligning the query with the finetuning distribution thus costs some semantic drift through the pivot. The gain is largest on MSD and Shakespeare, whose source registers are the most strongly marked, and negligible on ParaDetox, where source and target differ by a small lexical edit and little style remains for the normalizer to strip.

At inference time, relevant examples act on content rather than style: against random 5-shot they raise BLEU in all four domains (mean $+1.72$) but leave accuracy unchanged in three of four, reversing the pattern of Table~\ref{tab:exp_RAG}, where finetuning-time relevance was primarily an accuracy lever. Random examples are actively harmful---0-shot inference exceeds random 5-shot on BLEU in three of four domains. That 0-shot inference here also exceeds the 0-shot-\emph{finetuned} model of Table~\ref{tab:exp_RAG} on BLEU in three domains, neither prompt containing any example, indicates that the benefit of retrieval-augmented finetuning is retained in the weights.

\subsection{Experiments on Pretrained LLMs}\label{sec:exp_basellm}

To demonstrate the generalizability of RT-SFT, we repeated the experiment with three other pretrained LLMs under the best configuration of \textsection\ref{sec:exp_RAG}---relevant 5-shot examples with terminology and name retrieval (Table~\ref{tab:exp_basellm}). Style accuracy is again given by the per-domain BERT classifier of \textsection\ref{sec:method_acc}, and COPY provides the untransferred floor.

\begin{table*}[h]
    \centering
    \resizebox{\textwidth}{!}{%
    \begin{tabular}{@{}lcccccccc@{}}
        \toprule
        \multirow{2}{*}{\textbf{Pretrained LLMs}} & \multicolumn{2}{c}{\textbf{MSD}} & \multicolumn{2}{c}{\textbf{Shakespeare}} & \multicolumn{2}{c}{\textbf{GYAFC}} & \multicolumn{2}{c}{\textbf{ParaDetox}} \\
        \cmidrule(lr){2-3} \cmidrule(lr){4-5} \cmidrule(lr){6-7} \cmidrule(lr){8-9}
        & \textbf{Acc.} & \textbf{BLEU-ref} & \textbf{Acc.} & \textbf{BLEU-ref} & \textbf{Acc.} & \textbf{BLEU-ref} & \textbf{Acc.} & \textbf{BLEU-ref} \\
        \midrule
        COPY & 0.107 & 14.01 & 0.009 & 15.35 & 0.149 & 16.54 & 0.012 & 20.03 \\
        meta-llama/Llama-3.1-8B-Instruct & \textbf{0.763} & \textbf{13.79} & \textbf{0.842} & \textbf{17.35} & \textbf{0.612} & \textbf{24.17} & \textbf{0.912} & 19.91 \\
        gorilla-llm/gorilla-openfunctions-v2 & 0.661 & 11.07 & 0.733 & 17.11 & 0.573 & 21.15 & 0.874 & \textbf{19.93} \\
        mistralai/Mistral-7B & 0.692 & 12.21 & 0.749 & 15.97 & 0.552 & 22.23 & 0.806 & 18.45 \\
        facebook/opt-2.7b & 0.631 & 11.39 & 0.701 & 13.33 & 0.606 & 21.27 & 0.811 & 17.21 \\
        \bottomrule
    \end{tabular}%
    }
   \caption{TST finetuning performance across base LLMs (relevant 5-shot finetuning with terminology and name retrieval; synthetic corpus pooled from German and Chinese roundtrip pivots). The Llama-3.1-8B-Instruct row is the best configuration of Table~\ref{tab:exp_RAG}, repeated here for comparison. COPY returns the input unchanged and gives the untransferred floor for each domain's style classifier. Best result per column in bold, excluding COPY. All four base models exceed that floor in every domain; Llama-3.1-8B-Instruct leads on style accuracy in all four and on BLEU in three.}
    \label{tab:exp_basellm}
\end{table*}

Every model clears that floor in all sixteen cells, by margins of $0.403$ to $0.900$, so the synthetic corpus supplies a usable training signal irrespective of the base LLM. Llama-3.1-8B-Instruct is strongest on both metrics---leading style accuracy in all four domains and BLEU in three, with Gorilla-openfunctions-v2 ahead by $0.02$ on ParaDetox---and we therefore adopt it for \textsection\ref{sec:exp_RAG} and \textsection\ref{sec:exp_inference}. Gorilla and Mistral-7B are indistinguishable from each other on both metrics (average BLEU $17.32$ vs.\ $17.22$, accuracy $0.710$ vs.\ $0.700$). Scale is not the binding constraint: opt-2.7b, roughly a third the size of the others, trails Llama by only $0.095$ in average accuracy and still exceeds the 5-shot ICL baseline of Table~\ref{tab:exp_RAG} in three of four domains. As in \textsection\ref{sec:exp_RAG}, all four models fall below the input itself (COPY) on MSD and ParaDetox BLEU, where source and reference already overlap heavily.

\section{Conclusion}

We presented a method for Text Style Transfer that addresses the scarcity of parallel corpora by using roundtrip translation as a style normalizer to generate parallel training data. Because general-domain NMT preserves content while regressing toward generic phrasing, roundtrip-translating a monolingual in-style corpus yields the pseudo-parallel data that supervised finetuning requires, turning normalization from an inference-time patch into a data-generation tool. LoRA-finetuning an instruction-tuned LLM on this corpus improves style-classification accuracy over both In-Context Learning and Automatic Post-Editing in all four domains and in every configuration tested, and does so without either half of the method being sufficient alone: zero-shot RT-SFT, which uses no retrieval, already beats ICL---the same model and prompt without the synthetic corpus---and APE, the same corpus without the LLM stylizer. BLEU behaves as expected under the style--content trade-off \citep{Mukherjee2023,Pang2019}, and RT-SFT leads on both metrics in the two domains where source and reference genuinely diverge.

Retrieval augmentation applied at training time, rather than only in the prompt, improves style accuracy in all eight paired comparisons against randomly sampled examples, and its benefit is retained in the weights: a retrieval-augmented model prompted with no examples outperforms one finetuned without retrieval and likewise prompted with none. Terminology and name retrieval trades a small amount of accuracy for content preservation, and helps most where the target style is carried by a fixed vocabulary.

Roundtrip-translating the query at inference time to match the finetuning distribution raises style accuracy in all four domains and lowers BLEU in all four, a trade whose size tracks how strongly marked the source register is. Quantifying that trade-off under stylistically diverse and genuinely out-of-domain queries is the natural next step, along with varying the pivot language and extending a single RT-SFT-LLM to multi-style and non-binary transfer.

\section*{Limitations}
The main limitations of our work are as follows:
\begin{itemize} 
\item \textbf{Semantic drift and error propagation.} Our method relies on machine translation models to generate parallel finetuning datasets. As a result, its performance depends on the quality of the underlying NMT systems and their training data. We observed that when these models introduce errors or cause semantic drift during roundtrip translation, such inaccuracies become embedded in the synthetic parallel corpus used for finetuning. We applied post-processing steps to mitigate such effects, and further efforts could also be made to test various NMT methods or architectures to find the most suitable configuration for the TST task. These improvements and post-editing works, however, are beyond the scope of this study.

\item \textbf{Alternatives for the Current Roundtrip Translation Pipeline}. In this work, we primarily used Marian to train the NMT models and did not explore alternative methods or workflows for performing roundtrip translation. An intriguing potential alternative is to employ large language models to perform machine translation, either by ICL or finetuning, which might yield better results compared to the current Marian-based approach. However, we did not test these alternative approaches in the current study owing to constraints on time and space.

\item \textbf{Limits to domains with available corpus}. Due to data availability constraints, our experiments are conducted on four style domains, which may not fully capture the range of stylistic variations encountered in real-world scenarios. This limitation could introduce biases into our analysis and potentially restrict the generalizability of our methods. We selected benchmarks that are as diverse and distinctive as possible---spanning Shakespearean English, formality, expert--layman medical communication, and toxicity---in an effort to enhance the overall robustness and applicability of our method. We strive to enhance the generalizability of our experiments and demonstrate the effectiveness of our method in different domains and conditions.
\end{itemize} 

\bibliography{custom}

\clearpage

\appendix
\section{Prompt Templates}
\label{sec:appendix_prompt}

\textbf{TST finetuning prompts:} \\

We experimented with three potential prompt templates for RT-SFT (Table~\ref{tab_prompt_ft}). These prompts organize the query input sentence and several example sentence pairs into a prompt, with proper task descriptions and guidance for the generation. Templates (I) and (II) explicitly state the rewriting task, but have different orders of the example and query content. Template (III) is a classic Machine Translation prompt template with demonstrated effectiveness for Machine Translation with LLM. By changing language name to style domain name, we adapt it to guide LLM for text style transfer task. 

In this experiment (Table~\ref{tab:prompts_exp}), we conducted relevant 5-shot finetuning with terminology retrieval on Llama-3.1-8B-Instruct with the different templates, while leaving other conditions unchanged. Template (I) has the overall highest style accuracy score in the two tested domains. We thus used Template (I) throughout.

\begin{table}
\caption{Prompts for TST finetuning}
\setlength{\tabcolsep}{4pt}  
\small  
\begin{tabular}{p{0.2\columnwidth}p{0.7\columnwidth}}
\hline
\textbf{Prompt Template Index} & \textbf{Prompt Template Text} \\
\hline
I & Rewrite the following sentence into the style of [style name]. Here are [n] examples: Input: \textcolor{red}{[example input i]}. Output: \textcolor{red}{[example output i]}. Note that you may want to rewrite "[input term]" to "[output term]" for contextual consistency. Now go ahead: \textcolor{blue}{Input: [query input]}. The [style name] output: \\
\hline
II & Rewrite the following sentence into the style of [style name]: \textcolor{blue}{Input: [query input]}. Here are [n] examples: Input: \textcolor{red}{[example input i]}. Output: \textcolor{red}{[example output i]}. Note that word [input term] should be rewritten to [output term] for contextual consistency.\\
\hline
III & Note that word [input term] should be rewritten to [output term] for contextual consistency. General domain: \textcolor{red}{[example input i]}. [style name] domain: \textcolor{red}{[example output i]}. general domain: \textcolor{blue}{Input: [query input]}. [style name] domain: \\
\hline
\end{tabular}

\vspace{0.5em}
\footnotesize
\textbf{Candidate prompt templates for LLM style transfer finetuning}. 
\label{tab_prompt_ft}
\end{table}

\begin{table}[h!]
    \centering
    \resizebox{\columnwidth}{!}{%
    \begin{tabular}{@{}lcccc@{}}
        \toprule
        \multirow{2}{*}{\textbf{Template}} & \multicolumn{2}{c}{\textbf{MSD}} & \multicolumn{2}{c}{\textbf{Shakespeare}} \\
        \cmidrule(lr){2-3} \cmidrule(lr){4-5}
        & \textbf{Acc.} & \textbf{BLEU-ref} & \textbf{Acc.} & \textbf{BLEU-ref} \\
        COPY & 0.107 & \textbf{14.01} & 0.009 & 15.35 \\
        APE~\citeyearpar{9714400} & 0.672 & 12.21 & 0.693 & 11.90 \\
        Template I  & \textbf{0.763} & 13.79 & \textbf{0.842} & 17.35 \\
        Template II & 0.614 & 12.91 & 0.773 & 16.69 \\
        Template III& 0.721 & 13.63 & 0.751 & \textbf{18.01} \\
        \bottomrule
    \end{tabular}%
    }
    \caption{BLEU and acc. score across the MSD and Shakespeare domains for three potential templates. Template (I) has consistently higher style accuracy score, indicating robust ability in transferring to target style. Both Template (I) and (III) have stably high BLEU score.}
    \label{tab:prompts_exp}
\end{table}

\textbf{Terminology retrieval prompts:} \\
We retrieved domain-specific term and name pair lists for each domain to enhance TST performances, by calling the LLM twice for each instance in the synthetic parallel corpus. The prompts we used are shown in Table~\ref{tab_prompt_RAG}.

\begin{table}
\caption{Prompts for terminology RAG}
\setlength{\tabcolsep}{4pt}  
\small  
\begin{tabular}{p{0.2\columnwidth}p{0.7\columnwidth}}
\hline
\textbf{Prompt type} & \textbf{Prompt Text} \\
\hline
First round & Identify terminologies or character names in the sentence and return in comma separated format, without any additional explanation. Sentence: \textcolor{red}{[source-side sentence]}. Terminologies and names: \\
\hline
Second round & Find the counterpart of the word \textcolor{blue}{[source-side retrieved word]} in the following sentence and return a single word, without  any additional explanation. Sentence: \textcolor{red}{[target-side sentence]}:\\
\hline
\end{tabular}

\vspace{0.5em}
\footnotesize
\textbf{Prompts for terminology RAG}. The first prompt retrieves a list of terminologies and names from the source side sentence of each parallel instance, and for each of these retrieved words, the second prompt retrieves its counterpart in the corresponding target side sentence. 
\label{tab_prompt_RAG}
\end{table}

\section{Hyperparameters and experiment configurations}
\label{sec:appendix_config}

\textbf{LoRA finetuning hyperparameters:} \\
We set the learning rate to 2e-4, rank for the low-rank approximation is set to 512, the scaling factor is set to 256, and we use float16 data type. A dropout rate of 0.05 is applied. We save and evaluate the model every 2000 steps.\\
\\
\textbf{Marian Configurations:} \\
We used Marian for the Neural Machine Translation model pairs in the roundtrip translation system. In our system we used the Transformer architecture with R2L Reranking, with learning rate 0.0001, 49500 BPE operations, and step size 20000. 
\\

\textbf{Decoding configuration:} \\
All generation from the RT-SFT-LLMs uses greedy decoding with beam size 1 and
temperature 0. Generation is run with
vllm in
float16. The sketch-first retrieval pass of \textsection\ref{sec:method_RAG_ex_train}
is decoded with the same configuration, so the first-round sketch, and hence
the neighbour set it retrieves for the second round, is fixed for a given
query and checkpoint. 
\end{document}